\documentclass[conference]{IEEEtran}
\IEEEoverridecommandlockouts
\usepackage{cite}
\usepackage[T1]{fontenc}
\usepackage{amsmath,amssymb,amsfonts}
\usepackage{algorithmic}
\usepackage{graphicx}
\usepackage{textcomp}
\usepackage{xcolor}
\usepackage{amsmath}

\def\BibTeX{{\rm B\kern-.05em{\sc i\kern-.025em b}\kern-.08em
    T\kern-.1667em\lower.7ex\hbox{E}\kern-.125emX}}
\begin{document}

\title{A Comprehensive Comparison Between ANNs and KANs For Classifying EEG Alzheimer's Data\\

}

\author{\IEEEauthorblockN{Akshay Sunkara\IEEEauthorrefmark{1},
Sriram Sattiraju\IEEEauthorrefmark{1}, Aakarshan Kumar\IEEEauthorrefmark{1},
Zaryab Kanjiani\IEEEauthorrefmark{1} and Himesh Anumala\IEEEauthorrefmark{1}} 
\IEEEauthorblockA{\IEEEauthorrefmark{1}University of North Texas\\
Denton, United States of America\\ 
\{AkshaySunkara, SriramSattiraju, AakarshanKumar, ZaryabKanjiani, HimeshAnumala\}@my.unt.edu}}

\maketitle

\begin{abstract}
 Alzheimer's Disease is an incurable cognitive disease that affects thousands of people globally. While some diagnosis methods exist for Alzheimer's Disease, many of these methods can't catch Alzheimer's in it's earlier stages. Recently, researchers have considered the use of Electroencephalogram (EEG) technology for the diagnosis of Alzheimer's. EEG technology is a noninvasive method of recording the brain's electrical signals, and EEG data has shown differences between patients with and without Alzheimer's. In the past, Artificial Neural Networks have been used for the prediction of Alzheimer's from EEG data, but these prediction models can sometimes contain false positive diagnoses. In this study, we aim to compare losses between Artificial Neural Networks and Kolmogorov-Arnold Networks between multiple types of Epochs, Learning Rates, and Nodes. We have found that, throughout these different parameters, ANNs are more accurate in predicting Alzheimer's Disease from EEG signals.
\end{abstract}


\section{Introduction}
Alzheimer’s Disease (AD) is a neurocognitive disorder that gradually affects one’s cognitive capability. The prevalence of AD around the world is on the rise, with a predicted 152 million people to be affected by the disease in 2050 [1]. While AD is incurable, the early detection of AD has been found to help with managing cognitive symptoms and quality of life problems that might be related to the disorder [2]. AD is found to be most prevalent in people who are 65 years and older, which makes it hard to detect early on as diminishing cognitive capability can be otherwise attributed to old age within this group [3]. A solution that can aid with the early detection of AD for those of old age is crucial in ensuring that the highest quality treatment is possible for this group of people. 

The use of electroencephalography (EEG) systems for the recording of neurocognitive function and health has shown promising data when it comes to the detection of many neurocognitive disorders. EEG systems are a non-invasive method of recording a human brain’s electrical activity and have a large variety of use cases [4]. Previous studies have shown that EEG signals can serve as a potential marker for the early detection of neurocognitive disorders similar to AD [5]. Studies have also shown that AD patients with mild cognitive impairments have EEG signals that show abnormalities along many frequency bands in comparison to healthy patients [6].

Machine Learning (ML) systems also hold extreme promise when it comes to the detection of neurocognitive disorders through EEG signals. Previous studies have demonstrated that Artificial Neural Networks (ANN) are practical when it comes to classifying EEG signal data [7]. Along with this, studies have shown that the use of ANN’s are practical for the analysis and prediction of neurocognitive activity [8]. However, there is a lack of information on Kolmogorov-Arnold Networks (KAN) and their practicality for the classification/identification of EEG signal data for the detection of neurocognitive disorders and activity. KAN’s can potentially provide improvements in prediction when compared to their ANN counterparts, which is especially important when it comes to the early detection of AD for the mitigation of any false diagnoses [9]. The goal behind this study is to compare the effectiveness of KAN’s and ANN’s to predict EEG signals for the classification of neurocognitive disorders such as AD. Through parametric experiments that examined factors such as: epochs, learning rate, and how network structure affected the loss of both models, this study identified the strengths of each model in the analysis of EEG signals. 

\section{Overview of KAN's}
Compared to their traditional counterparts, ANNs, KANs are designed to be more efficient and accurate, handling fewer parameters and exhibiting a higher level of interpretability. The principal distinction between KANs and ANNs lies in the management of weights and activation functions. ANNs operate with fixed activation functions at their nodes and learnable weights along their edges, with these weights being modified during training. On the other hand, KANs use learnable activation functions along the edges, which are shaped as the network undergoes training. This enables KANs to display greater flexibility and adaptability while yielding higher accuracy. 

The Kolmogorov-Arnold Theorem, the foundation upon which KANs are based, asserts that if a multivariate continuous function \(f: [0, 1]^d \to \mathbb{R} \), then \(f\) can be expressed as a finite composition of continuous functions of a single variable and the binary operation of addition: \(g_{q}, \phi_{p, q}\) [10]. Formally, this is represented as:
\begin{equation}
f(x) = f(x_{1}, \ldots, x_{d}) = \sum_{q=0}^{2d} g_{q}(\sum_{p=1}^{d} \phi_{q, p}(x_{p}))
\end{equation}

This theorem is the core principle behind the training of KANs: each layer or node represents a continuous function of a single variable, while the overall network architecture combines these functions into a multivariate continuous function. This innovative approach utilizes simplicity to address complexities inherent in data.

\section{Methodology}

To prepare data for use with ANN's and KAN's, preprocessing steps were taken to ensure a clear classification standard was met for direct comparison between the two models.

\subsection{Alzheimer's Dataset Description}

The data used for this study was collected from an open source dataset of EEG readings from three different groups: Alzheimer’s Disease (AD), Frontotemporal Dementia (FD), and healthy [11]. These EEG readings were taken using a clinical EEG headset, and contains these features: Gender, Age, Health Classification, and MMSE (Mini-Mental State Examination) Score. Since this study is focused on the detection of Alzheimer's, EEG signals from those with FD were removed from the study, leaving a total of 65 people in the dataset. Out of all 65 people in the dataset, 36 were diagnosed with AD, and the remaining 29 were classified as healthy controls. The EEG data from this dataset consisted of signals recorded from 19 EEG scalp electrodes from participants in their resting state. 

Since this study is about the identification of AD through EEG signals, specific EEG scalp electrodes were chosen to best focus on areas of the brain that progressively change during AD. Studies had previously found that the hippocampus and amygdala were the primary areas that had been affected by AD [12]. This study analyzed the frontal cortex as well, as it primarily manages motor tasks, something that patients with AD can tend to lose control of over time [13]. For this study, 5 channels were chosen for analysis that most closely represented readings from these areas of the brain, such that areas of the brain affected by AD were most closely compared with healthy control subjects. 

\subsection{EEG Signal and Band Classification}

Once the appropriate EEG data was selected for this study from the dataset, pre-processing the EEG data was necessary for using in this study. Studies have shown that pre-processing EEG data is necessary for removing any excess noise that might be present from non-cerebral activity such that EEG signals can more closely represent cognitive activity [14] Processing the RAW EEG data started with a fifth order Butterworth band-pass filter from 0.3 to 25Hz, such that any excess noise from the EEG signals could be removed. This processing was implemented using the SciPy python library. After the data is processed through this filter, much of the excess noise is removed and the signals are much clearer in comparison to the original RAW EEG signals.

Along with signal filtering, EEG data can be categorized into multiple different bands. These different bands take place between the different frequencies of an EEG signal, and studies have shown that each of these bands all individually represent different characteristics of the brain [15]. For this study, we will be classifying processed EEG signals along four different bands: Theta (4-8 Hz), Alpha (8-13 Hz), Beta (13-30 Hz), Gamma (>30 Hz). 

\subsection{EEG Channel PSD}

A classification method is used for this EEG data such that each band can be easily analyzed for changes by our machine learning moels. For the case of this study, a mean Power Spectral Density (PSD) is performed on the EEG data such that each band in each channel has a mean value of power distribution in an EEG signal. The mean PSD was calculated from the RAW EEG data using the SciPy python library. Once all of these classification methods were applied to the EEG data, each subject in the dataset had a mean PSD value for each of the 5 channels and 4 bands, or a total of 20 mean PSD values per subject.

\subsection{Data Classification for Machine Learning}

Once the EEG data was classified, it was numericalized by mapping number values to the features and labels to pass it to our machine learning models. Additionally, we removed labels from the dataset such as:  'MMSE,' 'participant-id,' and 'File Name’ from the dataset to remove any prediction-irrevlevant data and created data loaders for training and testing with the remaining classes and gender labels.

For the design of the ANNs and KANs, we used 4 nodes in the input layer, a dropout rate of 0.5 to avoid overfitting, and 3 nodes in the output layer. We used CrossEntropyLoss() as the loss function and Adam as our optimizer. Our parametric experiment varied the number of epochs, learning rates, and nodes in the hidden layer to observe their effects on the losses.

\medskip

The data categorization process ensured that clear criterion were present for the classification of Alzheimer's Disease.

\section{Results and Discussion}
In our parametric experiments, we investigated the effects that several parameters: the number of epochs, the learning rate, and the number of nodes in a single hidden layer, had on the losses yielded by both networks. Additionally, we used the data retrieved from these experiments to determine the most optimal set of factors that led to the least loss. Based on this set of factors, we then trained both the models to plot confusion matrices to further compare both models, seen in Figure 1 and 2. The visualizations derived from these experiments offer valuable information that compare both the models’ strengths and weaknesses and the influence that certain factors had on the model’s loss.

 \begin{figure}[!t]
 \centering
 \includegraphics[width=3.5in]{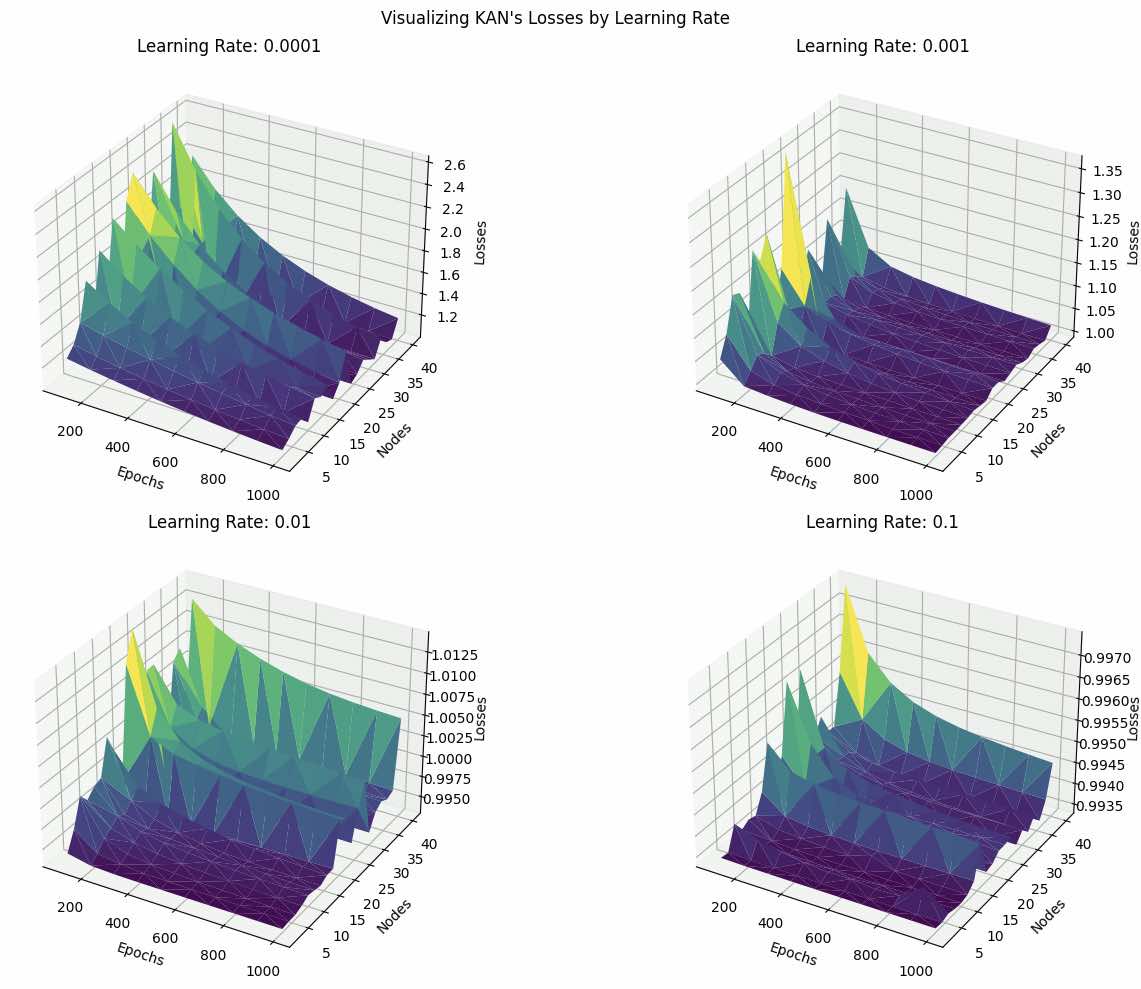}
 \caption{KAN's Losses by Learning Rate}
 \label{fig_kan}
 \end{figure}

 \begin{figure}[!t]
 \centering
 \includegraphics[width=3.5in]{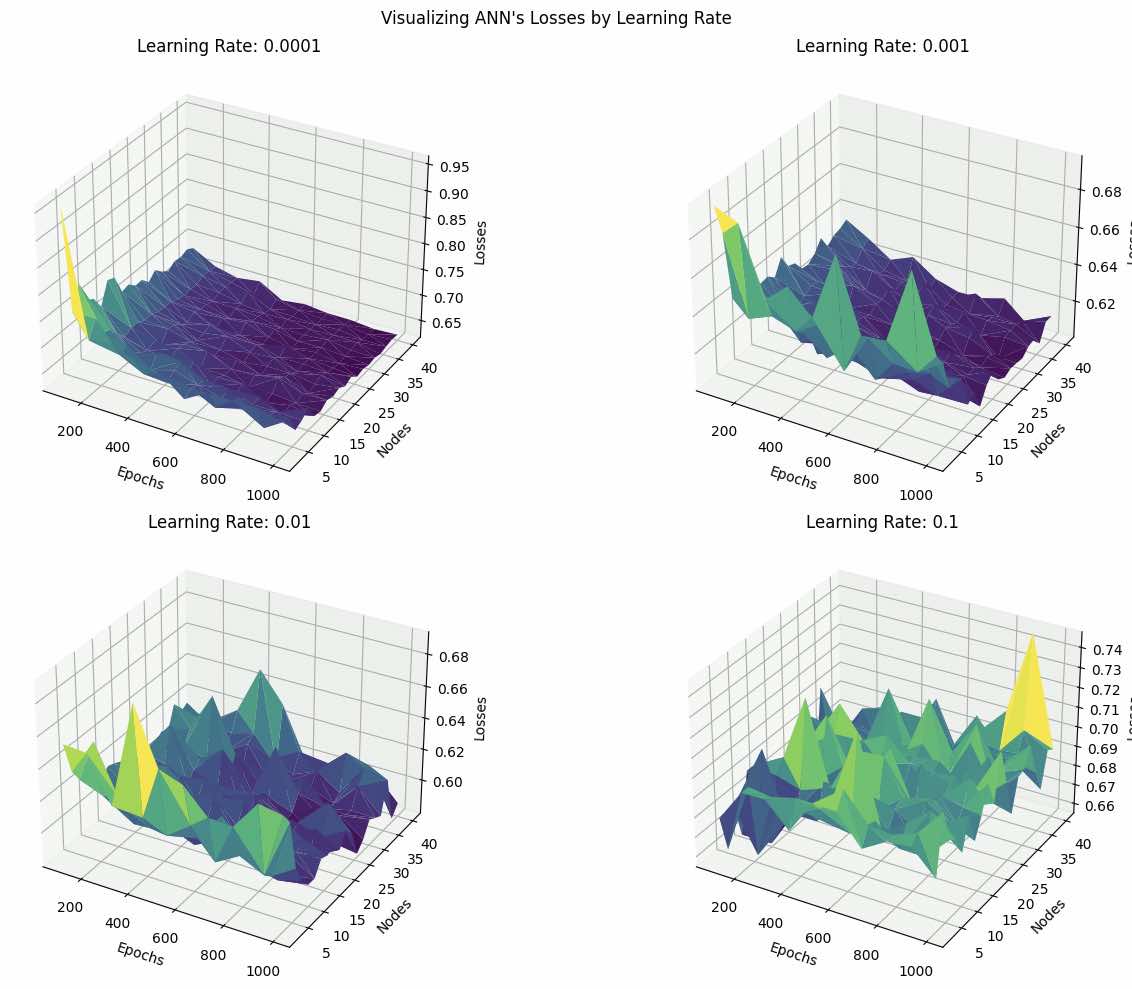}
 \caption{ANN's Losses by Learning Rate}
 \label{fig_ann}
 \end{figure}

\subsection{Performance}
In the analysis of Artificial Neural Networks (ANNs), distinct trends emerge across varying learning rates and network parameters. A learning rate of 0.001 consistently yields the lowest average loss, highlighting its efficacy in optimizing model performance. At a lower learning rate of 0.0001, both increasing the number of nodes and epochs contribute to a reduction in loss, with a pronounced decrease as epochs rise until reaching a plateau. Conversely, at a learning rate of 0.001, a significant reduction in loss is observed with an increase in the number of nodes, while the reduction attributed to increased epochs is less pronounced. Higher learning rates of 0.01 and 0.1 introduce greater variability in loss outcomes. Specifically, at 0.01, enhancements in both nodes and epochs are beneficial, whereas at 0.1, these increases result in the highest observed losses, indicating a potential overfitting or instability in the network at higher learning rates.

The behavior of Kernelized Associative Networks (KANs) with respect to learning rate adjustments shows a marked contrast to that of ANNs. KANs display less sensitivity to changes in the learning rate. Similar to ANNs, extending the number of epochs generally leads to lower or comparable losses. Notably, at a learning rate of 0.0001, KANs achieve lower losses with an increased number of nodes at fewer epochs, suggesting that simpler network architectures can be effective. At 0.001, both lower and higher node counts at fewer epochs result in similar losses, with a peak in losses observed at intermediate node levels. As the learning rate increases to 0.01, changes in epochs have minimal impact at lower levels but are more influential at higher epoch counts, whereas increasing node counts consistently results in higher losses. This trend persists at a learning rate of 0.1.

The analysis reveals that ANNs are significantly more influenced by learning rate adjustments than KANs, particularly in applications involving EEG data. The findings underscore the importance of selecting appropriate learning rates and network parameters to optimize performance for specific types of neural networks and datasets.

Furthermore, the data enabled us to determine the most optimal set of parameters that resulted in the lowest loss. For ANNs, the optimal parameters are [Epochs=1000, LR=0.01, Nodes=260], achieving a loss of 0.580. For KANs, the best set of parameters is [Epochs=1000, LR=0.1, Nodes=4], resulting in a loss of 0.993.

After training both models on these parameters, confusion matrices were plotted to further illustrate the advantages of both models (refer to Fig. 3 and Fig. 4).

\begin{figure}[!ht]
\centering
\includegraphics[width=3.5in]{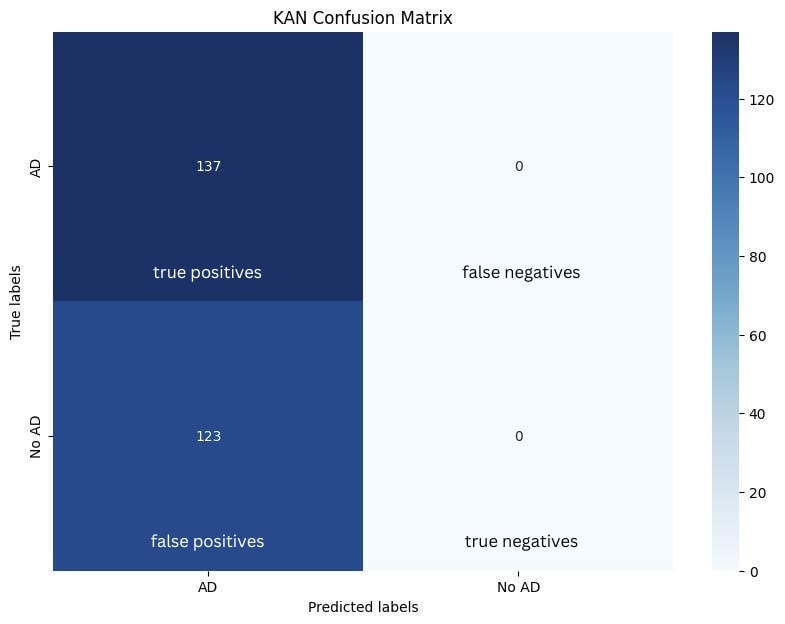}
\caption{KAN Confusion Matrix}
\label{fig_kan_conf}
\end{figure}

\begin{figure}[!ht]
\centering
\includegraphics[width=3.5in]{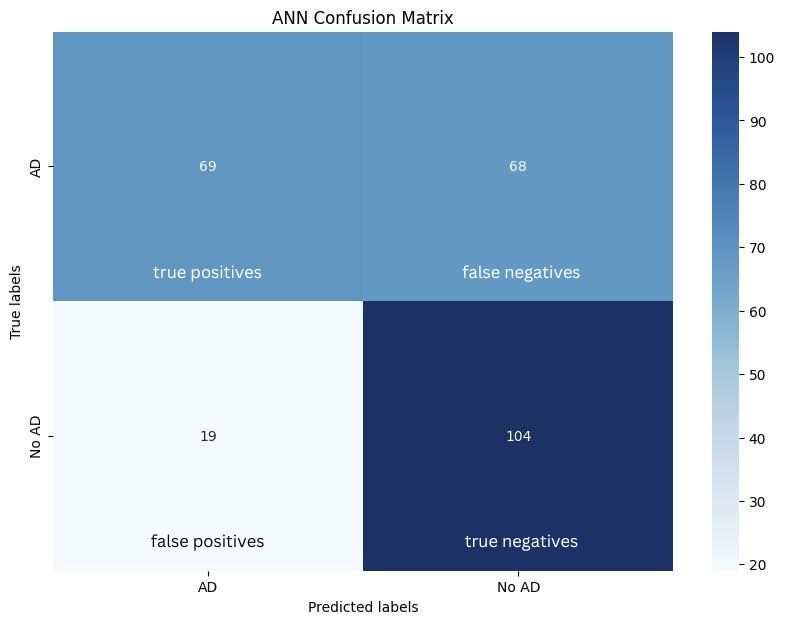}
\caption{ANN Confusion Matrix}
\label{fig_ann_conf}
\end{figure}

Along with these confusion matrices, an OLS Regression Plot (refer to Fig. 5 and Fig. 6) was made for each network to further compare the two models. 

\begin{figure}[!t]
\centering
\includegraphics[width=3.5in]{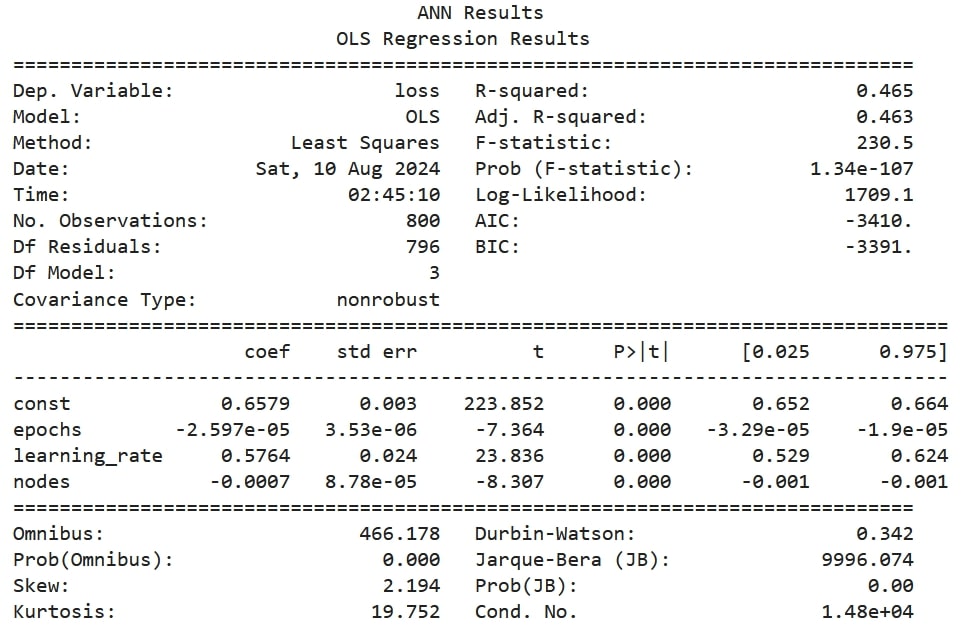}
\caption{ANN OLS Regression Results}
\label{fig_ann_conf}
\end{figure}

\begin{figure}[!t]
\centering
\includegraphics[width=3.5in]{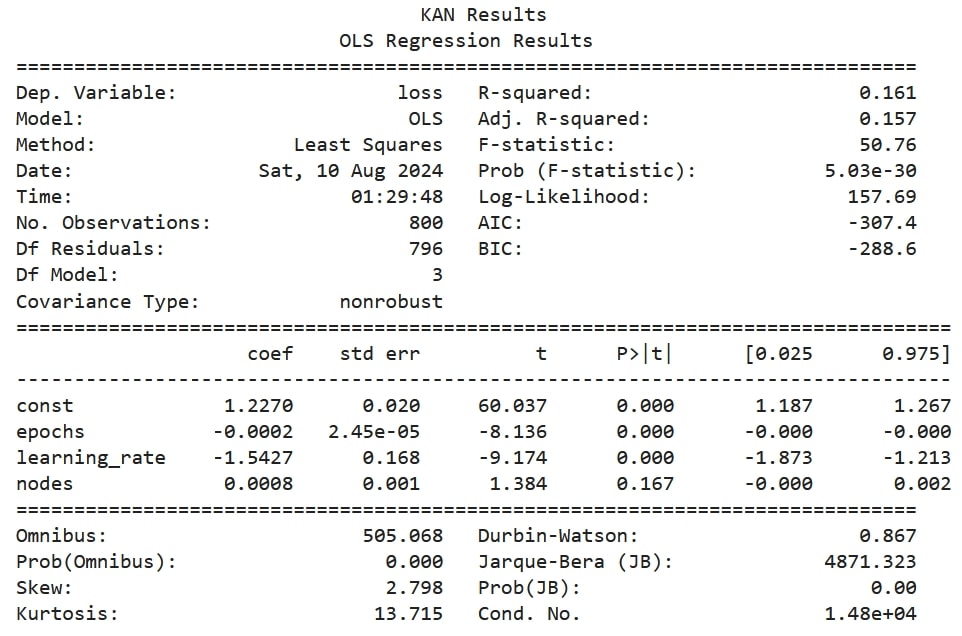}
\caption{KAN OLS Regression Results}
\label{fig_ann_conf}
\end{figure}

\medskip

These OLS regression plots also highlight the distinctions between ANNs and KANs: The R-squared (\(R^2\)) value for ANNs is 0.465, indicating that about 46.5\% of the variation in losses can be accounted for by the model’s inputs. In contrast, KANs have an \(R^2\) value of 0.161, suggesting that only about 16.1\% of the variation in losses is explained by the inputs. This likely implies that KANs are less dependent on factors such as epochs, learning rate, and the number of nodes. 

From these experiments, it appears that ANNs are more effective than KANs for classifying EEG data. However, it's important to note that our tests were limited to common parameters such as the number of nodes in the hidden layer, the training epochs, and the learning rate. Our OLS regression analysis indicates that KANs are less influenced by these factors, and many other parameters remain unexplored. It's possible that there are other settings for KANs that could lead to improved performance which we have yet to investigate.

\section{Conclusion}
In conclusion, this study analyzed the potential of KANs in predicting Alzheimer's Disease through EEG data. Through this study, we have found that KANs are simply not as accurate in predicting Alzheimer's Disease through EEG signals when compared to ANNs, having a much higher optimal loss when compared to KANs. The early detection of Alzheimer's Disease is necessary to ensure that the best treatment is met for people who are diagnosed with the disease. For a patient that is newly diagnosed with Alzheimer's, every aspect of their life changes, and the early detection of Alzheimer's makes the transition for these people to life with Alzheimer's much easier on them. In the future, expanding our research by training models across a broader range of parameters can be done to gain a deeper understanding of loss behaviors, such that we can understand what makes the best possible prediction model. Along with this, performing experiments on these two machine learning models with different types of data beyond EEG signals can potentially help increase prediction rates for both of these models, and is something that we will consider in the future.

\vspace{12pt}
\color{red}

\end{document}